\documentclass[conference]{IEEEtran}
\IEEEoverridecommandlockouts
\usepackage{cite}
\usepackage{algorithmicx}
\usepackage{CJKutf8}
\usepackage[ruled]{algorithm}
\usepackage{amsmath,amssymb,amsfonts}
\usepackage{graphicx}
\usepackage{subfigure}
\usepackage{booktabs}
\usepackage{textcomp}
\usepackage{xcolor}
\usepackage{algpseudocode}

\def\BibTeX{{\rm B\kern-.05em{\sc i\kern-.025em b}\kern-.08em
    T\kern-.1667em\lower.7ex\hbox{E}\kern-.125emX}}

\begin{document}
\begin{CJK}{UTF8}{gbsn}
\bstctlcite{IEEEexample:BSTcontrol}
    \title{A Dynamic Temporal Self-attention Graph Convolutional Network for Traffic Prediction}
  \author{Ruiyuan Jiang,
      Shangbo Wang,~\IEEEmembership{Member,~IEEE,} Yuli Zhang\\

  \thanks{}}

\markboth{IEEE TRANSACTIONS ON MICROWAVE THEORY AND TECHNIQUES, VOL.~60, NO.~12, DECEMBER~2012
}{Roberg \MakeLowercase{\textit{et al.}}: High-Efficiency Diode and Transistor Rectifiers}

\maketitle

\begin{abstract}
Accurate traffic prediction in real time plays an important role in Intelligent Transportation System (ITS) and travel navigation guidance. There have been many attempts to predict short-term traffic status which consider the spatial and temporal dependencies of traffic information such as temporal graph convolutional network (T-GCN) model and convolutional long short-term memory (Conv-LSTM) model. However, most existing methods use simple adjacent matrix consisting of 0 and 1 to capture the spatial dependence which can not meticulously describe the urban road network topological structure and the law of dynamic change with time. In order to tackle the problem, this paper proposes a dynamic temporal self-attention graph convolutional network (DT-SGN) model which considers the adjacent matrix as a trainable attention score matrix and adapts network parameters to different inputs. Specially, self-attention graph convolutional network (SGN) is chosen to capture the spatial dependence and the dynamic gated recurrent unit (Dynamic-GRU) is chosen to capture temporal dependence and learn dynamic changes of input data. Experiments demonstrate the superiority of our method over state-of-art model-driven model and data-driven models on real-world traffic datasets.  
\end{abstract}


%
\IEEEpeerreviewmaketitle


\section{Introduction}

\IEEEPARstart{W}{i}th the exponential growth of population, the usage of vehicles has increased significantly in the past few years. To better improve the efficiency of urban transportation networks, traffic prediction technique has received more and more attention which plays a key role in Intelligent Transportation System (ITS). In urban networks, traffic information such as flow, speed and density can be considered as a spatiotemporal data collected from multiple sensors. Making prediction on these time series serves as a foundation to many real-world applications \cite{2018Forecasting}. For example, online traffic prediction is critical in many applications such as vehicle navigation system \cite{Seungjae2006Short}, predictive bus control framework \cite{Matthias2017A} and travel time estimation. However, considering the complex temporal and spatial dependencies of traffic data, making accurate and reliable predictions has been a challenge task.\par
Most time series models such as Autoregressive Integrated Moving Average (ARIMA) \cite{2013ARIMA} and Gated Recurrent Unit (GRU) \cite{2014Empirical} mainly learn and preserve temporal correlations of traffic data. In contrast, some models were proposed to capture the spatial dependence to improve the model performance. For example, Graph Convolutional Network (GCN) \cite{0Traffic} can utilize spatial correlations among sensors through adjacency matrix. In addition, data collected by nearby loop sensors tend to have similar temporal characteristics. Hence, some hybrid spatiotemporal prediction models such as Temporal Graph Convolutional Network (T-GCN) \cite{2018T} and Traffic Graph Convolutional Long Short-term Memory (TGC-LSTM) \cite{0Traffic} network were developed to focus on the temporal and spatial dependencies of traffic data.\par
However, the existing traffic prediction methods have the following problems. Firstly, most methods capture the spatial correlations using a simple adjacency matrix. The adjacency matrix consists of 0 and 1, where 1 represents there is a connection between the two sensors where 0 denotes that there is no connection. So the adjacency matrix can only reflect the general network topology rather than accurately describe the spatial dependence between the tested sensors, which has a negative impact on the prediction accuracy. Secondly, most of the prevalent deep learning models perform inference in a static manner. That is, the model parameters are fixed after the training process, which may limit their representation power, efficiency and interpretability \cite{2016Adaptive, 2017Multi}. \par
To solve the above problems, we propose a dynamic temporal self-attention graph convolutional network (DT-SGN) in this paper which can describe the spatial correlation between selected sensors more accurately and make the model adaptable to different inputs. Inspired by the recent research on temporal graph neural network \cite{2018T} and dynamic neural networks \cite{2021Dynamic}, this model applies self-attention mechanism in T-GCN network and designs the framework of dynamic features to make the model parameters be dynamically rescaled according to the inputs to improve the efficiency. The main contributions of this paper are summarized as follows.\par
1) We proposed a novel self-attention GCN model to describe the topology structure of the urban network. That is, we calculate the correlation coefficient between the  selected road sections through multiplying the query and key of adjacency matrix, then use the coefficient matrix to replace the original adjacency matrix to further capture the spatial dependency of traffic data. Hence, the coefficient matrix can be considered as a trainable matrix which can be updated during model training process, so that the prediction performance can be improved.\par
2) We applied the attention mechanism on the output of GRU network for dynamic features, so that the spatial locations could be dynamically rescaled with attention to improve the representation power of the model. Moreover, the framework can adapt model parameters to different inputs which is effective in improving the representation power of networks with a minor increase of computational cost.\par
3) We tested the proposed model on two real-world traffic datasets, namely the SZ-taxi dataset and  Los-Loop dataset. We compare our proposed model with other recent state-of-art models through extensive experiments and demonstrate the superiority and accurate performance of our model.\par
The rest of the paper is organized as follows. In section \uppercase\expandafter{\romannumeral2}, we briefly review related work on model-driven and data-driven models for traffic prediction and the applications of attention mechanism in time series data prediction models. Section \uppercase\expandafter{\romannumeral3} introduces the DT-SGN framework and provides the methodology and formula derivation of our model. Section \uppercase\expandafter{\romannumeral4} provides the extensive numerical experimental results conducted on the two real-world datasets, followed by the conclusion and future work in section \uppercase\expandafter{\romannumeral5}.

\section{Related Work}
In this section, we review and summarize some related studies on modeling traffic data $X\in\mathbb{R}^{N\times{T}}$, where \emph{N} is the number of time series and \emph{T} is the number of time points. We will firstly introduce relevant work in traffic prediction. Then we will focus on the application of attention mechanism in time series prediction models.
\subsection {Traffic Prediction Models}
Various traffic prediction models have been investigated in existing literature which can be divided into two categories: the model-driven approaches and data-driven approaches. Firstly, the model-driven approaches mainly explain the instantaneous and steady-state relationships among traffic volume, speed and density \cite{8809901}. As one of the most popular model-driven models, ARIMA (Autoregressive Integrated Moving Average) based models such as S-ARIMA \cite{2003Modeling} and ST-ARIMA \cite{2018A} have raised great concern in traffic prediction. Besides ARIMA based models, the microscopic fundamental diagram model \cite{2013Impacts}, the cell transmission model \cite{2013Total} were also applied in traffic forecasting. However, the existing model-driven models cannot describe the variations of traffic data in real-world environments.\par
Secondly, data-driven models are used to predict and evaluate the traffic state based on the historical data \cite{2013Urban}\cite{2011Short}. In most recent years, novel deep learning-based traffic forecasting models such as deep bidirectional LSTM \cite{2018Deep} and shared hidden LSTM \cite{2016Deeptransport}. In addition, CNN based methods including \cite{2018Long}, \cite{2018Exploiting} and \cite{2019Deep} attempted to convert traffic status data into three-dimensional data, so that more effective features can be captured through novel models based on the CNN network. To better capture the temporal and spatial dependencies of traffic data, spatiotemporal prediction models such as GCN-DDGF \cite{2017Predicting}, TGC-LSTM \cite{0Traffic} and LSTM-GL-REMF \cite{2021Real} were proposed to utilize the temporal dependence in time series as well as capture the spatial correlation among sensors. Lin \emph{et al.} designed GCN-DDGF (Graph Convolutional Neural Network with Data-Driven Graph Filter) in order to learn the spatial dependence through GCN based framework while Cui \emph{et al.} proposed TGC-LSTM (Traffic Graph Convolutional Recurrent Neural Network) which defined spatial correlation based on traffic road network connectivity. In LSTM-GL-REMF, long short-term memory (LSTM) was chosen as the temporal regularizer to capture temporal dependency and Graph Laplacian (GL) served as the spatial regularizer to utilize spatial correlations to enhance the prediction performance.

\subsection {Attention Mechanism in Time Series Data Prediction}
Attention mechanism has been applied in time series data prediction combined with some deep learning models in recent researches. Kong \emph{et al.} \cite{2018HST} employed a hierarchical extension of the proposed ST-LSTM (HST-LSTM) in an encoder-decoder manner which modeled the contextual historic information in order to boost the prediction performance. Similarly, Zhang \emph{et al.} \cite{9488883} designed a Dual Attention-Based Federated Learning (FedDA) for wireless traffic prediction, by which a high-quality prediction model was trained collaboratively by multiple edge clients. In this framework, a dual attention scheme was proposed by aggregating the intra and inter-cluster models to construct the global model. Fu \emph{et al.} \cite{2021Hierarchical} proposed a spatiotemporal attention mechanism followed by graph convolutions to model the local correlations and patterns between traffic sensors on the same arterial road. Jin \emph{et al.} \cite{9713756} put forward GAN-Based Short-Term Link traffic prediction under a parallel learning framework (PL-WGAN) for urban networks which added spatial-temporal attention mechanism to adjust the importance of different temporal and spatial contexts. In addition, Duan \emph{et al.} \cite{9782553} applied fully dynamic self-attention spatiotemporal graph network (FDSA-STG) by improving the attention mechanism using graph attention networks (GATs). This model jointly modified the GATs and the self-attention mechanism that fully dynamically focused and integrated spatial, temporal and periodic correlations. Wang \emph{et al.} \cite{2021Spatio} proposed a novel spatial-temporal self-attention 3D network (STSANet) for video prediction, which integrated self-attention into 3D convolutional network to perceive contextual contents in semantic and spatiotemporal subspaces and narrows semantic and spatiotemporal gaps during saliency feature fusion. Chaabane \emph{et al.} \cite{2018Classification} used an adapted self-attention convolutional neural network to highlight the temporal evolution of land cover areas through the construction of a spatiotemporal map. This research proposed a new deep learning CNN based approach by introducing a self-attention mechanism that exploits both spatial and temporal dimension of image data. Kumar \emph{et al.} \cite{2020GCN} designed graph attention networks (GAT) to address the problem of treating all neighbors equally, which encoded the node of interest by using the weight features of its one-hot neighbors. Through it, local clustering coefficient solved the problem of unexplored intra-neighbor which can improve the model efficiency. 

\section{Methodology}
In this section, we will firstly give a problem description of this paper. Secondly, we introduce temporal graph convolutional (T-GCN) \cite{2018T} network which serves as a basis to understand our methodology. Then we will introduce the proposed dynamic temporal self-attention graph convolutional network (DT-SGN) model for traffic prediction task.

\subsection{Problem Description}
We assume a spatiotemporal setting for traffic data in this paper. In general, traffic data with \emph{N} sensors and \emph{T} time steps can be organized in a data matrix $X\in\mathbb{R}^{N\times{T}}$ in which each row and column corresponds to a sensor and a time step respectively. As shown in Figure 1, the goal of our research is to predict traffic information in a certain period of time based on the observed data achieved in previous time steps. In experiment section, we use traffic speed data as an example of traffic information.
\begin{figure}[H]
\centering
\includegraphics[width=0.45\textwidth, height=5.5cm]{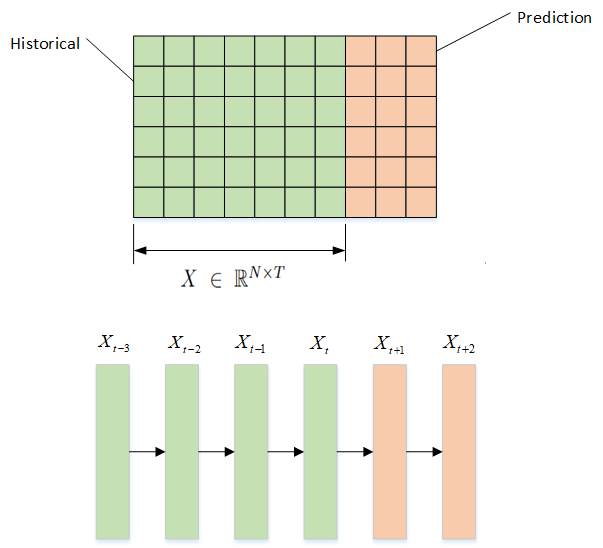}
\caption{\label{fig:1}Overview of traffic state prediction}
\end{figure}

\subsection{Temporal Graph Convolutional Network}
As mentioned, traffic data is a two-dimensional matrix of temporal series and spatial series. Thus, traffic prediction has always been a challenge task due to its complex spatial and temporal dependencies. Different from some classical prediction models which only capture the temporal dependence of traffic information such as ARIMA and LSTM, or other models which capture the spatial dependence such as GCN, T-GCN integrates GCN and GRU to predict future traffic data.\par
\begin{figure}[H]
\centering
\includegraphics[width=0.45\textwidth, height=4cm]{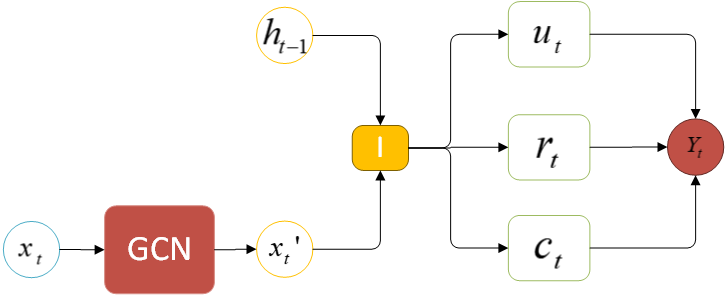}
\caption{\label{fig:2}Structure of T-GCN}
\end{figure}
Figure 2 shows the specific structure of a T-GCN cell, $h_{t-1}$ denotes the output at time \emph{t-1}, $u_t$, $r_t$ are update gate and reset gate at time \emph{t}. As shown in Figure 2, T-GCN firstly uses GCN network to learn spatial features of traffic data which can encode the topological structure of the road networks and the attributes on the roads to obtain the spatial dependence. Then the time series of spatial dependence are input into GRU network to capture the dynamic variation of traffic information on the roads for obtaining the temporal dependence. The output of GRU is eventually used as the prediction result.

\subsection{Dynamic Temporal Self-attention Graph Convolutional Network}
On the basis of T-GCN, we propose a novel dynamic temporal self-attention graph convolutional network (DT-SGN) for traffic prediction. The DT-SGN consists of two structures: self-attention graph convolutional network (SGN) and dynamic gated recurrent unit (Dynamic-GRU), where SGN can capture topological structure of urban networks and Dynamic-GRU is able to capture temporal dependence of the traffic data.\par
\emph{1) Self-attention graph neural network:}
To learn the spatial correlations of road sections, here we propose the self-attention graph convolutional network (SGN) model by employing a self-attention mechanism to model the spatial factor matrix. We use an adjacency matrix as the initial spatial score matrix in SGN, whose nodes represent the sensors, and whose edges represent dependency relationships between them. The adjacency matrix includes two values 0 and 1, where 1 denotes there is a connection between the two corresponding road sections and 0 denotes that there is no connection between them.\par
\begin{figure}[H]
\centering
\includegraphics[width=0.45\textwidth, height=5.5cm]{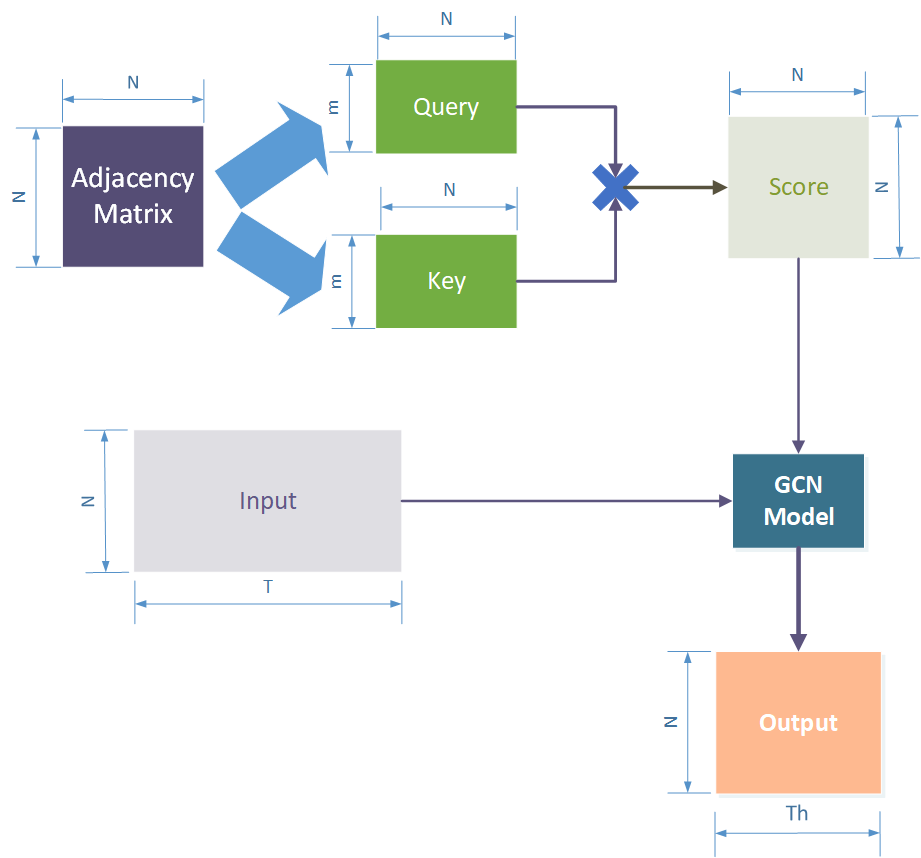}
\caption{\label{fig:3}Framework of SGN}
\end{figure}
Figure 3 presents the framework of SGN network which can be controlled through the query and key generated from the adjacency matrix. Unlike classical GCN model\cite{0Traffic}, SGN describes the correlation between road sections more accurately and extracts the road features according to the topological structure of the urban networks instead of directly using the adjacency matrix in a convolution layer. That is, based on the physical structure of sections connections, we scale up the correlation coefficients between sections with high spatial correlation while scale down the correlation coefficients between sections with low spatial correlation, so that the topological factors can be described as a matrix coefficients with a variety of values rather than the original matrix consisting of 0 and 1. To realize this, we use the concept of attention score to adjust the importance of different spatial contexts. The detailed methodology is shown as follows.\par
Given the adjacency matrix $A\in\mathbb{R}^{N\times{N}}$, where $A_{ij}$ represents the presence of a directed edge from section $i$ to section $j$. Two weight matrices $W_q$ and $W_k$ are allocated for the adjacency matrix to achieve the query and key value, as shown below:
\begin{equation}
Query = W_q{A}
\end{equation}
\begin{equation}
Key = W_k{A}
\end{equation}\par
Based on the query and key, we can extract the features of the road network structure through calculating the correlation of query and key. In this paper, we apply the attention mechanism to assess the correlation of the two matrices, as shown below:\par
\begin{equation}
S = Softmax(Query{Key^T})
\end{equation}
where $S$ represents the score matrix we use in model training, $Softmax(\cdot)$ is a normalized function.\par
Based on the score matrix and the traffic data matrix \emph{X}, we use the GCN model to construct a filter in the Fourier domain. Following the method proposed in \cite{2013Spectral}, we firstly estimate the matrix with added self-connections $\widetilde{S}$ and the degree matrix $\widetilde{D}$:
\begin{equation}
\widetilde{S} = \widetilde{S} + I_{N}
\end{equation}
\begin{equation}
\widetilde{D} = \sum_{j}{\widetilde{S_{i,j}}}
\end{equation}\par
where $I_{N}$ is a \emph{N} dimensional identity matrix.\par
Then the SGN model can be built by stacking multiple convolutional layers, which can be expressed as:
\begin{equation}
H^{n+1} = \sigma(\widetilde{D}^{-\frac{1}{2}}\widetilde{S}\widetilde{D}^{-\frac{1}{2}}H^{n}\theta^{n})    
\end{equation}
where $H^{n}$ is the output of the \emph{n}th layer, $H^{n+1}$ is the output of the \emph{n+1h}th lthayer, $\theta^{n}$ represents the model parameters of the \emph{n}th layer.\par
In this paper, we use two-layer GCN model \cite{2016Semi} to capture the spatial dependence of road networks, which can be expressed as:
\begin{equation}
S_{G} = \sigma(\widehat{S}ReLU(\widehat{S}XW_{1})W_{2})
\end{equation}
where $\widehat{S}=\widetilde{D}^{-\frac{1}{2}}\widetilde{S}\widetilde{D}^{-\frac{1}{2}}$ represents pre-processing step, $W_{1}$ is the weight matrix from input layer to hidden layer and $W_{2}$ is the weights matrix from hidden layer to output layer, $ReLU(\cdot)$ represents the Rectified Linear Unit, $S_{G}\in\mathbb{R}^{N\times{l}}$ denotes the output value with the prediction length \emph{l}.\par
In summary, we use the SGN model to capture the spatial dependence of traffic information. Attention score of adjacency matrix is used in model training process while we can update the score matrix with two weights $W_q$ and $W_k$ in each training epoch, so that the matrix can better capture the feature of road sections networks. The procedure of SGN is shown Algorithm 1.
\begin{algorithm}[H]
  \caption{Procedure of SGN}         
  \label{alg:Framwork}
  \begin{algorithmic}[1]
    \Require
      Traffic data matrix $X$; Adjacency matrix $A$;
    \Ensure
      Predicted value $S_{G}$;
    \State Initialize two weights $W_q$, $W_k$ for adjacency matrix.
    \State Initialize two weights $W_{1}$, $W_{2}$ for GCN network.  
    \State Set $epochs$ = $m$.
    \For{$i=0$ to $m$}
      \State $Query = W_q{A}$.
      \State $Key = W_k{A}$.
      \State $S = softmax(Query{Key^T})$.
      \State $\widehat{S}=\widetilde{D}^{-\frac{1}{2}}\widetilde{S}\widetilde{D}^{-\frac{1}{2}}$
      \State $S_{G} = \sigma(\widehat{S}ReLU(\widehat{S}XW_{1})W_{2})$.
      \State Update $W_q$, $W_k$, $W_{1}$ and $W_{2}$.
    \EndFor
  \\Get predicted value $S_{G}$ based on updated weights.
 \\\Return $S_{G}$;
  \end{algorithmic}
\end{algorithm}

\emph{2) Dynamic gated recurrent unit:}
To capture the temporal dependence of traffic information, we design a dynamic gated recurrent unit (Dynamic-GRU) in this paper. According to the temporal changes of traffic information, we dynamically rescale the features through adapting parameters to different inputs, which can effectively improve the representation 
power of networks with a minor increase of computational cost \cite{2021Dynamic} and improve the efficiency of the network. Figure 4 represents the concept of Dynamic-GRU network, where \emph{X} denotes the input of Dynamic-GRU network, \emph{O} represents the output of GRU network, \emph{Y} is the output of Dynamic-GRU. We apply attention $\alpha$ mechanism on the output of a GRU network, which is equivalent to performing computation with re-weighted parameters:
\begin{equation}
GRU(X, \Theta)\otimes{\alpha}=GRU(X, \Theta{\otimes{\alpha}})
\end{equation}
where \emph{GRU} denotes the process of GRU network, $\Theta$ is the GRU network parameters, $\otimes$ denotes the Kronecker product.
\begin{figure}[H]
\centering
\includegraphics[width=0.45\textwidth, height=4cm]{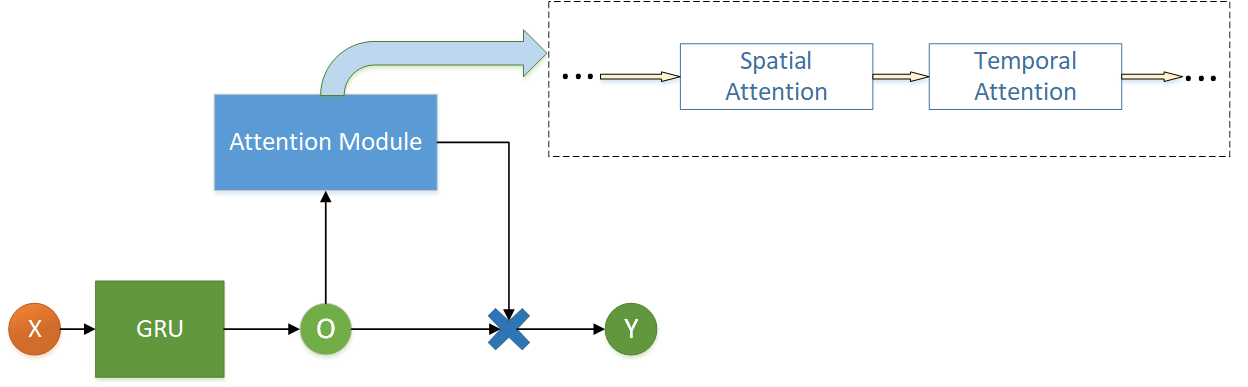}
\caption{\label{fig:4}Concept of Dynamic-GRU}
\end{figure}
The specific process is shown below. $X_{t}$ represents the input of GRU network at time \emph{t}, \emph{W} and \emph{b} represent the weights and bias in GRU training process, $h_{t-1}$ denotes the output at time \emph{t-1}, and $u_{t}$, $r_{t}$, $c_{t}$ are update gate, reset gate, cell gate at time \emph{t}, and $h_{t}$ is the output at time \emph{t}.
\begin{equation}
u_{t} = \sigma(W_{u}[X_{t}, h_{t-1}]+b_{u})
\end{equation}
\begin{equation}
r_{t} = \sigma(W_{r}[X_{t}, h_{t-1}]+b_{r})
\end{equation}
\begin{equation}
c_{t} = tanh(W_{c}[X_{t}, (r_{t}*h_{t-1})]+b_{c})
\end{equation}
\begin{equation}
h_{t} = u_{t}*h_{t-1}+(1-u_{t})*c_{t}
\end{equation}\par
Then, we add an attention layer on the output of hidden layer to rescale the features according to the input of the network. Instead of using pooling operations to efficiently gather global information in attention mechanism, we adopt convolutions to encode local information. Based on the hidden state $H = {h_{1}, h_{2},\cdots, h_{n}}$ obtained from GRU network, the weights of each characteristic $\alpha_{i}$ can be obtained through the similarity of query and key value of hidden state:
\begin{equation}
\alpha_{i} = Softmax(\frac{Q_{h}K_{h}}{\lvert Q_{h} \rvert \lvert K_{h} \rvert})
\end{equation}
where $Q_{h}$ and $K_{h}$ are query and key value of hidden state.\par
\begin{figure*}[t]
\centering
\includegraphics[width=0.9\textwidth, height=7cm]{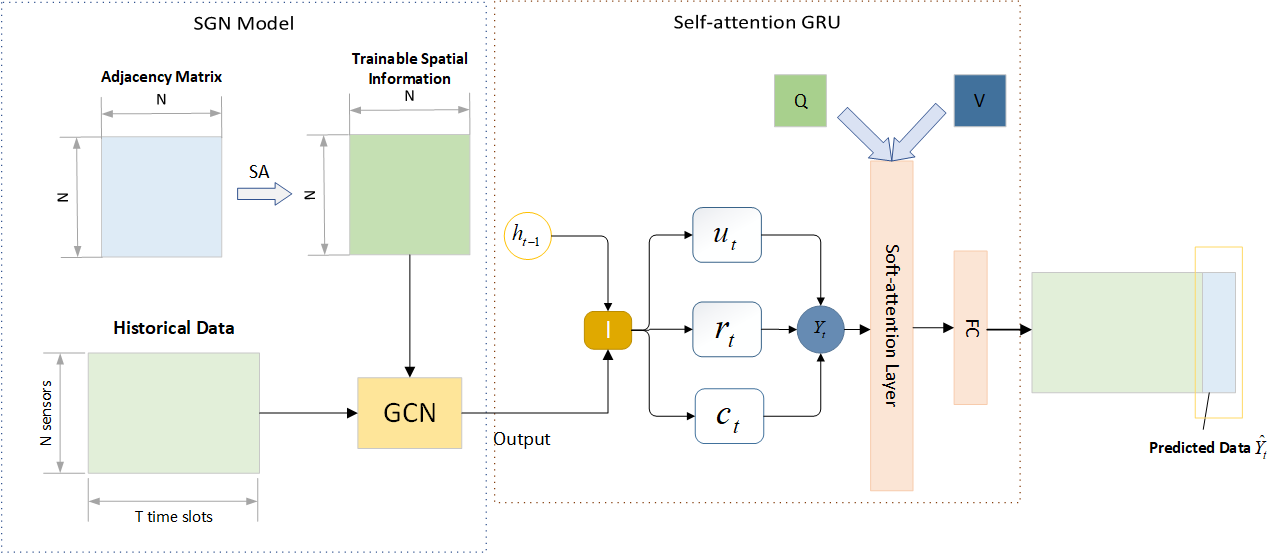}
\caption{\label{fig:5}DT-SGN framework}
\end{figure*}
And the attention function can be designed, the calculation of context containing local information is:
\begin{equation}
C_{t} = \sum_{i=1}^n{\alpha_{i}*h_{t}}
\end{equation}
Finally, we add a full-connected layer on the context and calculate the output of Dynamic-GRU network:
\begin{equation}
Y = W_{G}C_{t}+b_{G}
\end{equation}
where $W_{G}$ and $b_{G}$ are weight and bias allocated for full-connected layer. 

\emph{3) Dynamic temporal self-attention graph neural network:}
According to the temporal and spatial correlation of traffic information, we propose the dynamic temporal self-attention graph neural network (DT-SGN) in this research. The proposed model joins SGN and Dynamic-GRU in series to capture the spatial and temporal dependencies of traffic information simultaneously. As shown in Figure 5, the left side is the framework of SGN model and the right side is the framework of Dynamic-GRU model, where SA represents the self-attention score process, $h_{t-1}$ denotes the output at time \emph{t-1}, $FC$ denotes the full-connected layer. Based on the framework shown in Figure 5, the specific process of the DT-SGN model is shown as below.\par
According to the traffic data matrix $X$ and the adjacency matrix $A$, we firstly apply SGN model to learn the spatial dependence of traffic information:
\begin{equation}
G_t = SGN(X_{t}+\varepsilon_{i, t})    
\end{equation}
where $G_t$ represents the output of SGN network at time \emph{t}, $SGN(\cdot)$ is the process of SGN network, $X_{t}\in\mathbb{R}^{N\times{i}}$ is used to present the traffic data on each section at time \emph{i}, $\varepsilon_{i, t}$ is a zero-mean noise-term.\par
Then we use the output of SGN as the input of Dynamic-GRU network, as shown below:
\begin{equation}
\widehat{Y_{t}}=DGRU(G_t)0
\end{equation}
where $\widehat{Y_{t}}$ represents the predicted output of Dynamic-GRU network, $DGRU(\cdot)$ denotes the process of Dynamic-GRU.\par
In the model training process, we update the model parameters in the structures to minimize the error between real traffic data and predicted value. The loss function of our network can be shown as below.
\begin{equation}0
L = \lvert \lvert Y_{t}-\widehat{Y_{t}} \rvert \rvert+\lambda L_{reg}
\end{equation}
where $Y_{t}$ represents the real traffic data, $\widehat{Y_{t}}$ is the predicted value, $\lambda$ is a hyper-parameter and $L_{reg}$ is the L2 regularization term that helps to avoid an overfitting problem. In summary, the procedure of DT-SGN is shown in Algorithm 2.
\begin{algorithm}[H]
  \caption{Procedure of DT-SGN}         
  \label{alg:Framwork}
  \begin{algorithmic}[1]
    \Require0
      Traffic data matrix $X$; Adjacency matrix $A$;
    \Ensure
      Predicted value $Y_{t}$;
    \State Initialize model parameters $\theta_s$ for SGN network.
    \State Initialize model parameters $\theta_g$ for Dynamic-GRU net- 
    \State work, and $W_G$, $b_G$ for the full connected layer. 
    \State Set $epochs$ = $n$.
    \For{$i=0$ to $n$}
      \State Train SGN (Algorithm 1) on data X, and capture
      \State spatial dependence as shown in Eq. (7) and get $G_{t}$.
      \State $h_{t}=GRU(G_t)$ as shown in Eq. (9) to Eq. (12).
      \State $C_{t} = att(h_{t})$ as shown in Eq. (13) and Eq. (14).
      \State $Y_{t} = FC(C_{t})$ as shown in Eq. (15).
      \State $L = \lvert \lvert Y_{t}-\widehat{Y_{t}} \rvert \rvert+\lambda L_{reg}$.
      \State Update $\theta_s$, $\theta_g$, $W_G$ and $b_G$.
    \EndFor
  \\Get predicted value $Y_{t}$ based on updated parameters.
 \\\Return $Y_{t}$;
  \end{algorithmic}
\end{algorithm}

\section{Experiments}
\subsection{Data Description}
To demonstrate the superiority of our proposed model, we evaluate the performance of DT-SGN on two datasets: SZ-taxi dataset and Los-loop dataset.\par
(1) SZ-taxi dataset: This dataset consists taxi trajectory data of Shenzhen from Jan. 1 to Jan. 31, 2015. The study area includes 156 road sections of Luohu District. This experimental dataset mainly consists of two parts. The first part is a two-dimensional matrix which describes the speed data of 156 selected road sections over time, where each row denotes one road and each column is the traffic speed on the roads in different time periods. We apply 15 minutes as the time interval between each time period. The second part is an $156\times{156}$ adjacency matrix which describes the basic topological structure of road network. Each row represents one road and the values in the matrix denotes the connectivity between the roads, where 1 represents the two roads are connected and 0 denotes the opposing situation.\par
(2) Los-loop dataset: This dataset consists traffic speed data collected from 207 sensors in Los Angles County from Mar. 1 to Mar. 7, 2012. Similarly, the data consists of two parts, the first part is a two-dimensional matrix which describes the speed data collected from the selected sensors and the second part is an adjacency matrix which represents the connectivity in road networks. We apply 5 minutes as the time interval between each time period.\par 
In the experiments, we use 80 percent of the data in the dataset as the training set and remaining data is used as the test data. In addition, the input data is normalized to the interval [0, 1] for the convenience of the calculation process.

\subsection{Evaluation Metrics}\par
We use five metrics to evaluate the prediction performance of the DT-SGN network: Root Mean Square Error (RMSE), Mean Absolute Error (MAE), Accuracy (ACC), Coefficient of Determination ($R^2$), Explained Variance Score (\emph{var}). Assume $y_{i,j}$ and $y^{pred}_{i,j}$ represent the real traffic speed and the predicted value of the j\emph{th} time instant in the i\emph{th} road; $Y$ and $Y^{pred}$ represent the the set of $y_{i,j}$ and $y^{pred}_{i,j}$ respectively, $\bar{Y}$ is the average of $Y$; $M$ is the number of sampling time instants, $N$ is the number of roads. The metrics are respectively, computed by the following equations:\par
(1) Root Mean Square Error (RMSE):
\begin{equation}
RMSE = \sqrt{\frac{1}{MN}\cdot\sum_{j=1}^M{\sum_{i=1}^N{{(y_{i,j}-y^{pred}_{i,j})^2}}}}
\end{equation}\par
(2) Mean Absolute Error (MAE):
\begin{equation}
MAE = \frac{1}{MN}\cdot\sum_{j=1}^M{\sum_{i=1}^N{\lvert y_{i,j}- y^{pred}_{i,j}\rvert}}
\end{equation}\par
(3) Accuracy ($ACC$):
\begin{equation}
ACC = 1-\frac{\lvert\lvert y_{i,j}- y^{pred}_{i,j} \rvert \rvert_F}{\lvert\lvert y_{i,j}\rvert \rvert_F}
\end{equation}\par
(4) Coefficient of Determination ($R^2$):
\begin{equation}
R^2 = 1-\frac{\sum_{j=1}^M{\sum_{i=1}^N{{(y_{i,j}-y^{pred}_{i,j})^2}}}}{\sum_{j=1}^M{\sum_{i=1}^N{{(y_{i,j}-\bar{Y})    ^2}}}}
\end{equation}\par
(5) Explained Variance Score (\emph{var}):
\begin{equation}
VAR = 1-\frac{f_{var}(Y-Y^{pred})}{f_{var}(Y)}
\end{equation}\par
where $f_{var}$ refers to the variance function.

\subsection{Baselines}
We compare the performance of DT-SGN with the following baseline methods: \par
(1) Attention Temporal Graph Convolutional Network (A3T-GCN): As referred in \cite{ijgi10070485}, A3T-GCN captures the temporal dependence of data using GRU and learns the spatial dependence of data through GCN. Moreover, a self-attention is introduced to adjust the importance of different time points and assemble global temporal information;\par
(2) Temporal Graph Convolutional Network (T-GCN): As referred in \cite{2018T}, T-GCN applies GCN and GRU integrated to capture the temporal and spatial dependencies of traffic speed data simultaneously;\par
(3) Long Short-Term Memory (LSTM): LSTM has a strong ability to model time series data and normally has better prediction performance than linear models and shallow-learning models.\par
(4) Bidirectional Gated Recurrent Unit (Bi-GRU): As shown in \cite{9485098}, similar to the use of LSTM, Bi-GRU can be seen as a structure that combines the forget and output gates which specializes in exploring temporal dynamics.\par
(5) Autoregressive Integrated Moving Average (ARIMA): As referred in \cite{2013ARIMA}, ARIMA is a conventional time series model;\par

\subsection{Model Parameters}\par
In this paper, we use the same datasets as referred in \cite{2018T}. In \cite{2018T}, a series of experiments have been conducted on the hyper-parameter of prediction models. Hence, we refer to the experiment results to set the model parameters in our paper. We choose the following model parameters in experiments: learning rate, batch size, training epoch and the number of GRU unit. In the experiment, we manually adjust and set the learning rate to 0.001, the batch size to 33, the training epoch to 200, the number of GRU unit to 32, as shown in Table 1:
\begin{table}[!ht]
\centering
\setlength{\tabcolsep}{5mm}{
\renewcommand\arraystretch{1.5}{
\caption{}
\begin{tabular}{c|c}\hline
Parameters & Value \\ \hline
learning rate &  0.001\\ \hline
batch size & 33\\\hline
training epoch & 200\\ \hline
GRU unit & 32\\\hline
\end{tabular}
}
}
\end{table}

\subsection{Modeling Results and Discussions}\par
To demonstrate the superiority of DT-SGN network, we show the prediction results of our model and compare the prediction0 performance00 of DT-SGN network with the selected baselines on two real-world datasets: SZ-taxi dataset and Los-loop dataset.\par
\begin{figure}[H]
\centering
\subfigure[]{   
\includegraphics[width=0.45\textwidth, height=1.8cm]{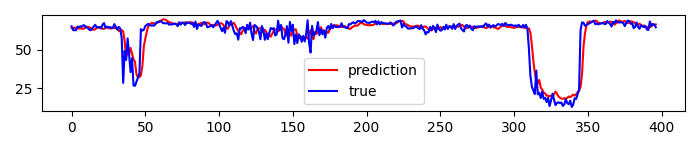}
}
\subfigure[]{
\includegraphics[width=0.45\textwidth, height=1.8cm]{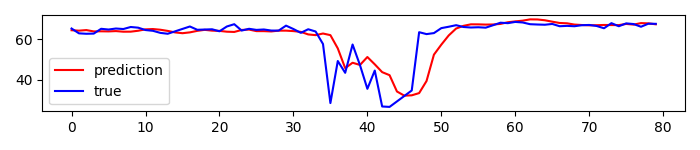}
}
\subfigure[]{
\includegraphics[width=0.45\textwidth, height=1.8cm]{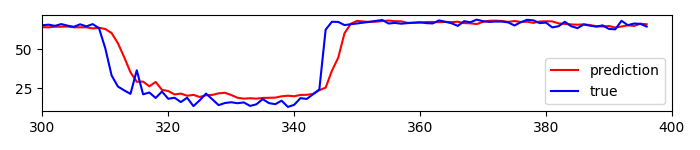}
}
\caption{\label{fig:6}Prediction performance of DT-SGN network for prediction horizon of 15 minutes on Los-loop dataset. (a) is the prediction performance of whole test time, (b) is the prediction performance from time 0 to time 80, (c) is the prediction performance from time 300 to time 400.}
\end{figure}
\emph{1) Los-loop dataset:}
Figure 6 shows the prediction result of DT-SGN model on Los-loop dataset, where the red curve represents the prediction results, the blue curve denotes the real data. From the overall situation shown in Figure 6 (a), it can be seen that the prediction results are in good agreement with the real data, which verifies the efficiency of DT-SGN network. Also, the curve of the prediction results is smoother than the curve of real data. It is mainly because that the GCN model uses a smooth filter to capture the spatial dependence through moving the filter constantly, which leads to a smaller change in the prediction results compared with the real data. From the detailed situation of Figure 6 (a), we find that the prediction accuracy is high when the data is steady, while the accuracy becomes lower when there is a sudden increase or decrease of data. To get better sense of the phenomenon, we zoom in on two phases in the test set that contain significant data fluctuations which are shown in Figure 6 (b) and Figure 6 (c), where we find that the differences between predicted value and the real data get larger where there is a sharp change in data. The main reason is that our proposed model is a data-driven model which predicts the future data based on historical data patterns. Hence, the prediction results is more like a regular smooth curve, so that when there is a large change in data, the prediction results will be delayed, leading to the low prediction accuracy.

\begin{figure*}[t]
\centering
\subfigure[]{
\includegraphics[width=0.45\textwidth, height=6cm]{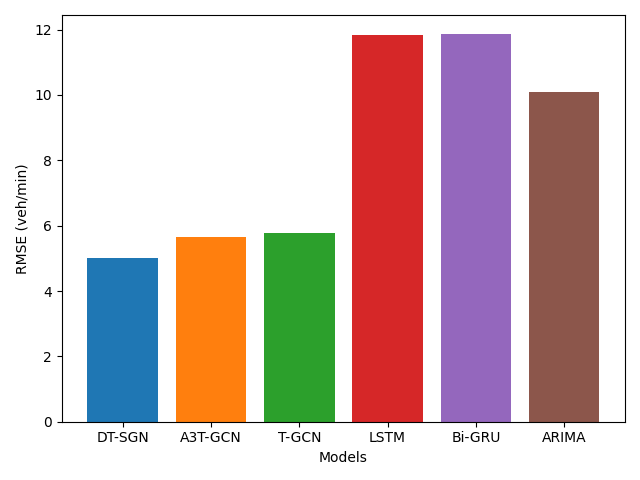}
}
\subfigure[]{
\includegraphics[width=0.45\textwidth, height=6cm]{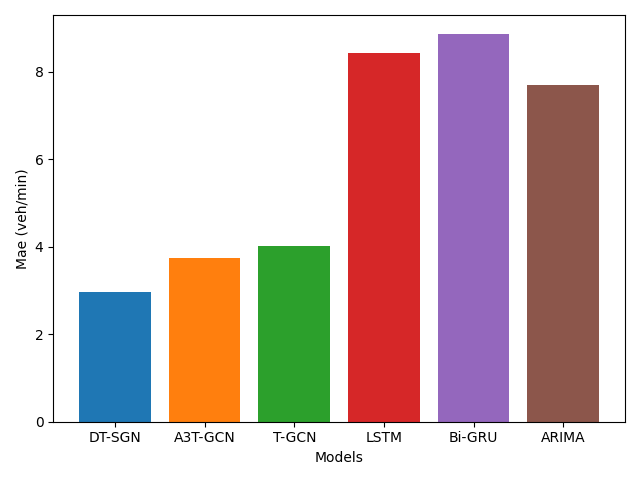}
}
\caption{\label{fig:6}Prediction capacity on Los-loop dataset. (a) is the RMSE of tested methods and (b) is the MAE of tested methods.}
\end{figure*}

\begin{table*}[!ht]
\centering
\setlength{\tabcolsep}{5mm}{
\renewcommand\arraystretch{1.5}{
\caption{}
\begin{tabular}{c||c|c|c|c|c|c}\hline
Metric & DT-SGN & A3T-GCN & T-GCN & LSTM & Bi-GRU & ARIMA\\ \hline
$ACC$ & \textbf{0.9146} & 0.9036 & 0.9016 & 0.7987 & 0.7983 & 0.8271\\ \hline
$R^2$ & \textbf{0.8679} & 0.8318 & 0.8244 & 0.2652 & 0.2624 & 0.0214\\ \hline
$VAR$ & \textbf{0.8684} & 0.8322 & 0.8260 & 0.2654 & 0.2854 & 0.002\\\hline
\end{tabular}
}
}
\end{table*}
Then we compare the prediction performance of the DT-SGN model and baseline methods. Figure 7 shows the prediction performance given by RMSE and MAE, it can be seen that the DT-SGN network obtains the best prediction performance among all prediction methods, proving the effectiveness of spatiotemporal traffic prediction. Among all data-driven models, the RMSE error and MAE error of DT-SGN network are significantly lower than the other models, while prediction performances of A3T-GCN and T-GCN rank second and third respectively, which are far ahead of other models. Also, compared with ARIMA which is a model-driven model, DT-GCN demonstrates the obvious superiority according to the two metrics as well. \par
To further compare the prediction efficiency of DT-SGN model and baselines, we apply accuracy, coefficient of determination and explained variance score to evaluate the prediction performance of prediction models. As shown in Table \uppercase\expandafter{\romannumeral2}, the value of the three evaluation metrics of our proposed model remain the highest compared with the baselines. For example, $ACC$ of DT-SGN are improved by approximately 1$\%$, 14.5$\%$, 10.6$\%$ compared with A3T-GCN, LSTM and ARIMA and $R^2$ of DT-SGN are improved by approximately 4.5$\%$, 5.2$\%$, 227.2$\%$ compared with A3T-GCN, T-GCN and LSTM. Also, compared with A3T-GCN and T-GCN, $VAR$ of DT-SGN are increased by 4.3$\%$ and 5.2$\%$. Moreover, the superiority of T-GCN based models including DT-SGN, A3T-GCN and T-GCN is mainly due to the fact that the framework of such models capture the spatial and temporal dependencies of traffic information which can improve the prediction performance.

\emph{2) Sz-taxi dataset:}
Figure 8 shows the prediction result of DT-SGN model on Sz-taxi dataset, where the red curve represents the prediction results, the blue curve denotes the real data. From the blue curve in Figure 8 (a), it can be seen that the data of Sz-taxi dataset takes the form of constant oscillations, which shows high frequency and amplitude of change across the whole time line. Compared the prediction results with the real data, we find that the two curves are basically consistent, while the curve of prediction results generally appears below the curve of real data. To evaluate the prediction performance of DT-SGN in more detail, we take two pieces of data from the test set and compare them with the prediction value, as shown in Figure 8 (b) and Figure 8 (c). Figure 8 (b) shows the prediction results and real data from time instant 0 to time instant 80 while Figure 8 (c) shows the prediction results and real data from time instant 300 to time instant 400. We find that compared to the frequency and amplitude of oscillation of real data, the two elements of prediction results significantly decrease, so that the curve of prediction results is smoother. The reason of this phenomenon is similar to that we show in the above which is caused by the smooth filter of GCN network.\par

\begin{figure}[http]
\centering
\subfigure[]{   
\includegraphics[width=0.45\textwidth, height=1.9cm]{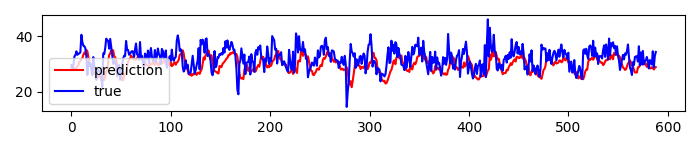}
}
\subfigure[]{
\includegraphics[width=0.45\textwidth, height=1.9cm]{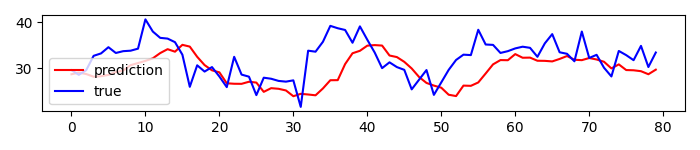}
}
\subfigure[]{
\includegraphics[width=0.45\textwidth, height=1.9cm]{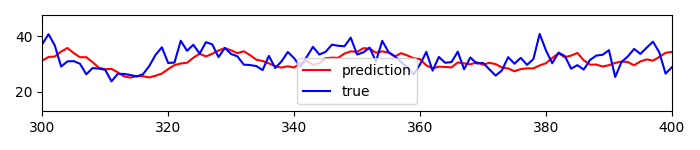}
}
\caption{\label{fig:6}Prediction performance of DT-SGN model for prediction horizon of 15 minutes on Sz-taxi dataset. (a) is the prediction performance of whole test time, (b) is the prediction performance from time 0 to time 80, (c) is the prediction performance from time 300 to time 400.}
\end{figure}

Figure 9 shows the prediction performance of DT-SGN and baseline methods given by RMSE and MAE on Sz-taxi dataset. It can be seen that the RMSE error and MAE error of DT-SGN network are the smallest among all methods. The prediction performance of DT-SGN is slightly better than that of Bi-GRU while RMSE and MAE of DT-SGN are 4.3997 and 2.9236, RMSE and MAE of Bi-GRU are 4.4108 and 3.1507. Similarly, DT-SGN is superior to A3T-GCN and T-GCN by the two metrics and the superiority becomes more obvious compared with LSTM and ARIMA. That is, our model shows superiority over the data-driven methods and model-driven methods o0f selected baselines.\par
Again, we evaluate the performance of DT-SGN and baselines by accuracy, coefficient of determination and explained variance score. The three evaluation metrics also proves the superiority of DT-SGN as shown in Table \uppercase\expandafter{\romannumeral3}. For example, $ACC$ of DT-SGN are improved by approximately 15.4$\%$, 12.3$\%$ and 27.1$\%$ compared with A3T-GCN, T-GCN and LSTM  while $R^2$ of DT-SGN are approximately 17.7$\%$, 14.1$\%$, 23.8$\%$ higher than that of A3T-GCN, T-GCN and LSTM. Also, compared with A3T-GCN and T-GCN and LSTM, $VAR$ of DT-SGN are increased by 17.9$\%$, 14.4$\%$ and 34.7$\%$. In addition, as a model-driven model, the prediction performance of ARIMA is relatively lower than data-driven models such as Bi-GRU and LSTM. This is mainly because that ARIMA model has difficulty in processing the long-term time series data. Also, ARIMA model is calculated by the error of each node and averaging, so when the data fluctuate widely, the accuracy of ARIMA model is not satisfactory.

\begin{figure*}[t]
\centering
\subfigure{
\includegraphics[width=0.45\textwidth, height=6cm]{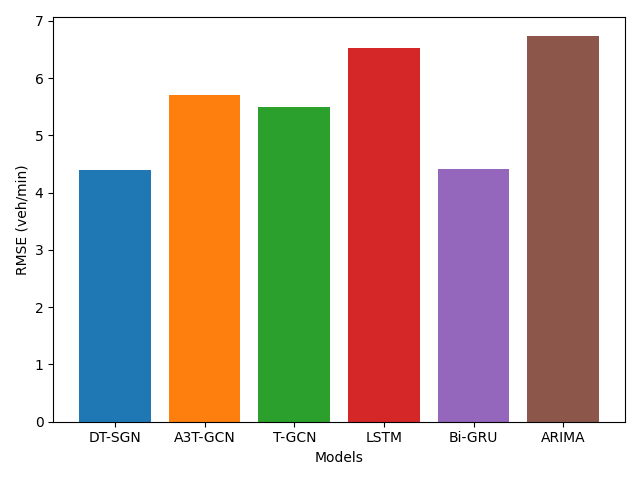}
}
\subfigure{
\includegraphics[width=0.45\textwidth, height=6cm]{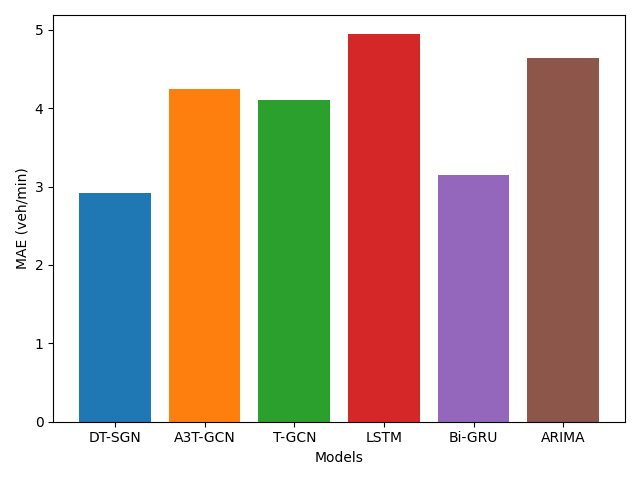}
}
\caption{\label{fig:6}Prediction capacity on Los-loop dataset. (a) is the RMSE of tested methods and (b) is the MAE of tested methods.}
\end{figure*}

\begin{table*}[!ht]
\centering
\setlength{\tabcolsep}{5mm}{
\renewcommand\arraystretch{1.5}{
\caption{}
\begin{tabular}{c||c|c|c|c|c|c}\hline
Metric & DT-SGN & A3T-GCN & T-GCN & LSTM & Bi-GRU & ARIMA\\ \hline
$ACC$ & \textbf{0.6935} & 0.6023 & 0.6174 & 0.5455 & 0.6927 & 0.3833\\ \hline
$R^2$ & \textbf{0.8225} & 0.7011 & 0.7232 & 0.6095 & 0.82161\ & 0.04673\\ \hline
$VAR$ & \textbf{0.8272} & 0.7012 & 0.7233 & 0.6119 &  0.8275 & 0.0141\\ \hline
\end{tabular}
}
}
\end{table*}

\emph{3) Analysis:} Firstly, the experiment results proves the high prediction precision and the ability in capturing temporal and spatial dependencies of traffic information of DT-SGN. As shown in Figure7 and Figure 9, DT-SGN demonstrates the superiority compared with the baseline methods according to RMSE and MAE. It can be seen that compared with data-driven models and model-driven model, DT-SGN obtains the best prediction performance among all methods. Also, Table \uppercase\expandafter{\romannumeral2} and Table\uppercase\expandafter{\romannumeral3} indicate the high prediction efficiency of DT-SGN by accuracy, coefficient of determination and explained variance score compared with the most existing prediction methods.\par 
Also, we find that the prediction performance of the same prediction methods exists a little differences according to the distribution of data. The reason of differences in prediction accuracy is shown as below. DT-SGN and the most baselines are data-driven models which obtain the data patterns through model training, so that the accuracy will decrease when the data appears great changes or the data fluctuates frequently. The overall distribution of data in Los-loop data set is stable except for two sharp declines while the data in Sz-taxi dataset constantly fluctuates in a wide range. Compared with the data of Sz-taxi dataset, the overall distribution of data in Los-loop data set is stable except for two sharp declines, so the prediction accuracy is high during the period of steady flow and get affected when there is a large change in the data. Comparatively speaking, the accuracy of prediction results tested on Sz-taxi dataset is lower than that tested on Los-loop dataset caused by the persistent unstable data series.

\section{Conclusion}
In this paper, we propose a dynamic temporal self-attention graph convolutional (DT-SGN) network to predict traffic information. Firstly, we propose a self-attention GCN (SGN) model which applies self-attention mechanism to model the adjacency matrix to make the topological structure of urban networks a trainable matrix during a deep learning model training process, aiming at better capturing the spatial dependence of traffic data. In addition to SGN operation, we add an attention layer on the output of GRU network for dynamic features, so that the framework can adapt model parameters to different inputs, which can improve the prediction performance with a minor increase of computation cost. These technical developments together make the model beyond the state-of-art methods of traffic prediction. Also, developing dynamic features with attention mechanism can improve the representation power of the model.\par
The proposed model has been implemented and intensively evaluated in two real-world datasets: Los-loop dataset and Sz-taxi dataset. The computational experiments demonstrate the superiority in traffic state prediction. Compared with the baseline methods including A3T-GCN, T-GCN, LSTM, Bi-GRU and ARIMA, the proposed model provides the best prediction performance according to different kinds of evaluation metrics. Future work is to integrate the model for traffic signal control or autonomous vehicle control. The idea aims to firstly use DT-SGN network to predict future traffic state which can be applied as the basic of data for control systems. In the long run, our network can be further developed and combined with \emph{Q} network or deep \emph{Q} learning for traffic management to improve the existing reinforcement learning model.


\bibliographystyle{IEEEtran}
\bibliography{reference}
\end{CJK}
\end{document}